\documentclass[lettersize,journal]{IEEEtran}
\usepackage{amsmath,amsfonts}
\usepackage{algorithmic}
\usepackage{algorithm}
\usepackage{array}
\usepackage[caption=false,font=normalsize,labelfont=sf,textfont=sf]{subfig}
\usepackage{textcomp}
\usepackage{stfloats}
\usepackage{url}
\usepackage{verbatim}
\usepackage{graphicx}
\usepackage{cite}
\usepackage{multirow}
\usepackage{color}
\usepackage{booktabs}
\begin{document}

\title{Distortion‐Resilient Robotic Imitation Learning for Autonomous Cable Routing}


\author{Hao Wang, Fu-Zhao Ou, Shiqi Wang$^{*}$,~\IEEEmembership{Senior Member,~IEEE,} Zhaolin Wan$^{*}$, and Xiaopeng Fan,~\IEEEmembership{Senior Member,~IEEE}
\thanks{Hao Wang is with the School of Artificial Intelligence, Harbin Institute of Technology, Harbin, Heilongjiang 150001, China, and also with the Department of Computer Science, City University of Hong Kong, Kowloon, Hong Kong SAR 999077, China (E-mail: ho.wong@my.cityu.edu.hk).}
\thanks{Fu-Zhao Ou and Shiqi Wang are with the Department of Computer Science, City University of Hong Kong, Kowloon, Hong Kong SAR 999077, China (E-mail: fuzhao.ou@my.cityu.edu.hk; shiqwang@cityu.edu.hk).}
\thanks{Zhaolin Wan is with the School of Artificial Intelligence, Harbin Institute of Technology, Harbin, Heilongjiang 150001, China (E-mail: zlwan@hit.edu.cn).}
\thanks{Xiaopeng Fan is with the School of Artificial Intelligence, Harbin Institute of Technology, Harbin, Heilongjiang 150001, China, also with the Pengcheng Laboratory, Shenzhen, Guangdong 518055, China, and also with the Suzhou Research Institute, Harbin Institute of Technology, Suzhou, Jiangsu 215104, China (E-mail: fxp@hit.edu.cn).}
\thanks{$^{*}$Corresponding Authors: Zhaolin Wan, Shiqi Wang}
}


\markboth{Journal of \LaTeX\ Class Files,~Vol.~14, No.~8, August~2021}%
{Shell \MakeLowercase{\textit{et al.}}: A Sample Article Using IEEEtran.cls for IEEE Journals}


\maketitle

\begin{abstract}
The rapid development of intelligent control methodologies has endowed robots with powerful autonomous intelligence. Cable routing, a ubiquitous foundational task in industry, provides a rigorous benchmark for robotic dexterity and sequential decision-making. In these practical scenarios, image observation distortion frequently occurs. Samples characterized by low-quality image observations often hinder accurate model training, posing challenges to the reliability and accuracy of intelligent control systems. Nevertheless, no dedicated intelligent control solution has been proposed for scenarios of image signal distortion. Meanwhile, image quality information has not been sufficiently exploited to further enhance the performance of intelligent control methodologies. To this end, we propose a novel robotic imitation learning framework that comprises an image quality assessment module, a confidence-based learning mechanism, and a decision-making module, which is designed to maintain high performance even under distorted image observations. In the proposed framework, the image quality assessment module synergizes with the confidence-based learning mechanism to enhance the efficacy of the decision-making module. Specifically, the image quality assessment module is incorporated to extract image quality information from image observations, while the confidence-based learning mechanism adaptively prioritizes challenging samples to improve learning effectiveness. The decision-making module determines appropriate discrete skills or continuous actions. Experimental results demonstrate that our formulated framework enhances the overall performance of the decision-making module.
\end{abstract}


\section{Introduction}
\IEEEPARstart{T}{th} field of robotics is poised for a promising future, with significant advances paving the way for innovative applications such as in child care \cite{yang2019integrating}, service \cite{jeong2017task, liu2021scene}, gait generation \cite{rosa2022topological}, manipulation \cite{sun2022motion}, healthcare \cite{zeng2024design}, industry \cite{campagna2024promoting}, navigation \cite{chen2024toward, li2024neuronsgym, hu2024efficient, liu2025toward}, and cognition \cite{gimenez2023you}. One of the primary reasons for the rapid development of robotics is that vision-based imitation learning (IL) methodologies equip robots with powerful autonomous intelligence capabilities \cite{luo2024multi, zhang2024one, shi2024ranking, lin2025learning}. IL is a specialized area within machine learning in which autonomous agents acquire the ability to interact effectively with environments by observing and mimicking human actions \cite{zare2024survey}. It typically conceptualizes behavior acquisition as a supervised learning task and achieves high computational efficiency by leveraging the extensive body of research on supervised learning models within the machine learning domain \cite{pomerleau1991efficient, ross2011reduction}. This methodology is particularly advantageous in situations where explicit programming of the desired behavior is either difficult or impractical \cite{schaal1999imitation}. It drives autonomous agents to generalize from demonstrated behaviors, enabling adaptation to new and complex environments \cite{osa2018algorithmic}. Therefore, IL approaches not only improve the efficiency and accuracy of robotic operations but also significantly broaden the scope of applications for autonomous systems. 

In practical robotic implementations, particularly with remote-deployed robots and fast-moving cameras, image distortions commonly degrade the fidelity of visual observations, thereby undermining the performance of vision‐driven decision‐making systems \cite{steffens2021robustness, li2024motion, ryoo2024image}. However, to date, no IL paradigm has been explicitly designed to compensate for such distortions during autonomous decision-making processes, despite the substantial advances in learning‐from‐demonstration methods for robotic control \cite{zare2024survey, mahmoudi2024leveraging}. Meanwhile, although extensively employed to gauge image fidelity, image quality assessment (IQA) methodologies remain unexplored within the domain of robotic intelligent control \cite{ryoo2024image, zhang2024embodied}. Importantly, the prevailing IQA techniques are chiefly grounded in human‐visual‐system (HVS) modeling, and their extension to robotic perception and decision‐making contexts is still in its infancy \cite{jiang2020blind, ou2022novel, song2022medical, ou2023troubleshooting, gao2024blind, ryoo2024image, zhang2024embodied, ou2024refining, ou2024clib, hu2024spatio}. These specific challenges underscore the critical need for robust image processing algorithms and adaptive learning techniques to enhance the reliability and efficiency of robotic systems. Addressing these issues is essential for the seamless integration of robots into complex environments, ultimately fostering the development of more autonomous and intelligent robotic solutions.

\begin{figure}[tbp]
	\centering
	\includegraphics[width=\linewidth]{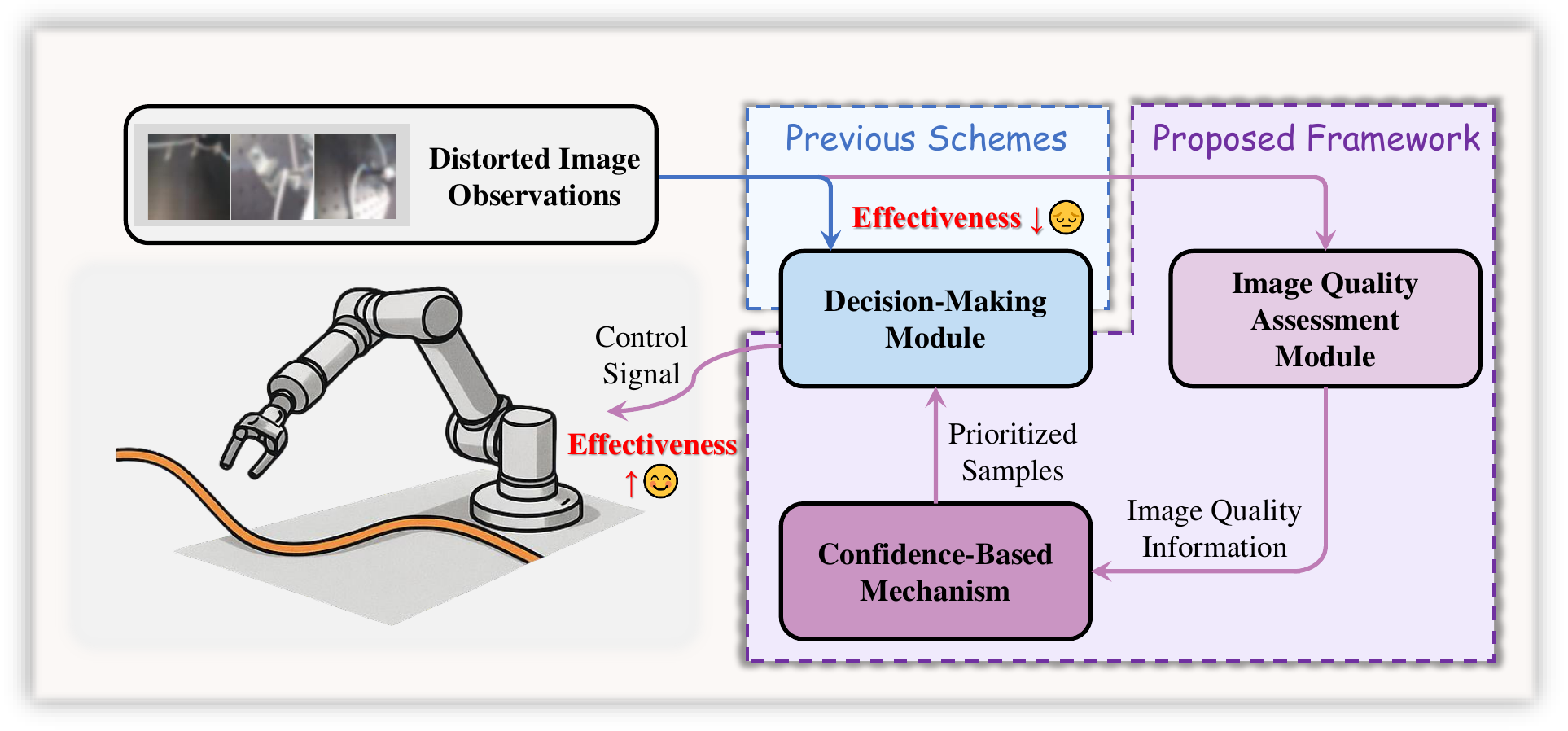}
	\caption{Comparison of the proposed robotic IL framework with previous schemes. Samples with low-quality image observations are typically challenging to learn accurately in previous schemes. In our proposed framework, leveraging decision-making-related quality information extracted by the IQA module, the confidence-based learning mechanism dynamically emphasizes challenging samples during training. Based on the integration of the IQA module and the confidence-based learning mechanism, the learning process of the decision-making module is systematically guided to ensure the effective acquisition of capabilities.}
	\label{framework}
\end{figure}

The cable routing task is a quintessential work in the field of robotics \cite{luo2024multi}. It exemplifies the difficulties found in intricate multistage robotic manipulation scenarios: managing deformable objects, integrating visual perception feedback, and executing extended sequences of actions that must all be performed correctly to accomplish the overall task \cite{luo2024multi}. The successful completion of this task will have profound implications for the advancement of robotic deployment in industrial scenarios \cite{luo2024multi}. In this study, we propose a robust intelligent control framework for the cable routing task, which improves the reliability and effectiveness of autonomous robotic systems even when image signals are interfered with. Moreover, with the intention of directly establishing a relationship between image distortions and robot visual system (RVS), we also construct the MyRMB dataset to quantify i\underline{M}age qualit\underline{y} via \underline{R}obotic \underline{M}anipulation \underline{B}ehaviors. Concurrently, we develop an objective IQA method to quantify the quality of visual observations, furnishing robotic decision‑making systems with essential prior knowledge. Specifically, as depicted in Fig. \ref{framework}, this proposed framework incorporates an IQA module and a confidence-based learning mechanism to improve the effectiveness of the decision-making module. The IQA module is utilized to derive decision-making-related visual quality information from input observations, while the confidence-based learning mechanism dynamically prioritizes samples according to their training difficulty. Experimental results suggest that our proposed framework exhibits excellent performance in the presence of image distortion. Robotic decision‑making systems achieve enhanced decision accuracy across a broad spectrum of visual fidelity conditions. Hence, our framework provides a new solution for addressing intelligent control issues in environments afflicted by image signal interference.

The major contributions of this paper are as follows:
\begin{itemize}
    \item We propose a novel robotic IL framework that orchestrates the synergy between an IQA module and a confidence-based learning mechanism to optimize the effectiveness of the decision-making component, which can achieve superior performance in scenarios with image distortions. 
	\item We first construct a dataset that assesses image quality via robotic manipulation behaviors. Furthermore, we also develop an objective IQA method to measure the quality of image observations captured by robot cameras.
	\item Our study pioneers the sufficient exploitation of the quality information extracted by the IQA methodology in the intelligent control domain, which offers a novel problem-solving perspective for the application of autonomous agents in image distortion scenarios and the development of intelligent control in autonomous systems.
\end{itemize}

\section{Related Works}
\label{Related Works}
In this section, we review the relevant literature in the fields of robotic intelligent control and IQA.

\subsection{Intelligent Control in Robotics}
Intelligent control methodologies have emerged as the predominant paradigm in robotic automation studies. Finn \textit{et al.} \cite{finn2016unsupervised} and Levine \textit{et al.} \cite{levine2018learning} proposed learning-based methodologies for enhancing robotic interactions, with the former focusing on an unsupervised approach for predicting object motion through action-conditioned video prediction and the latter emphasizing the application of deep learning for hand-eye coordination in robotic grasping. To further address the long-horizon planning issues, hierarchical frameworks are frequently utilized. For example, Luo \textit{et al.} \cite{luo2024multi} delved into the intricacies of mastering multistage tasks in robotic manipulation, which underscores the significance of employing hierarchical frameworks and judiciously selecting controllers as pivotal strategies to proficiently manage setbacks and flaws caused by low-level execution. Wang \textit{et al.} \cite{wang2024geneworker} introduced a novel robotic reinforcement learning (RL) algorithm that employs a collaborative framework between a generator network and a worker network to learn flexible neural skills, enabling effective decision-making in complex environments. In recent years, emerging techniques, such as large language models (LLMs) and Transformers, have also been integrated into robotic algorithms. In SayCan \cite{ahn2022can}, LLMs are leveraged to empower robots to carry out textual instructions in real-world scenarios, showcasing improved performance via grounding language in affordances. Robotics Transformers \cite{brohan2022rt, brohan2023rt, padalkar2023open} are a series of scalable machine learning models engineered specifically for real-world robotic control operations, leveraging large-scale, heterogeneous, and task-agnostic datasets to attain superior performance in specific downstream tasks.

Nevertheless, existing methodologies are not optimized for distorted image observations. Furthermore, they did not consider using image quality information as prior knowledge to effectively enhance the performance of intelligent control methods. In contrast, our study introduces a novel framework that leverages an IQA module to obtain quality priors relevant to robotic decision-making, facilitating autonomous agents' accurate learning from distorted visual observations in intelligent control tasks.

\subsection{Image Quality Assessment}
In the literature, most current IQA methods focus on leveraging deep learning models to map the perceptual quality of visual content to quality scores. For instance, in MetaIQA \cite{zhu2020metaiqa}, meta-learning \cite{vanschoren2018meta} is introduced to learn the shared prior knowledge from a range of distortions, enabling quick adaptation to unseen distortion types. HyperIQA \cite{su2020blindly} adopts a self-adaptive hyper network architecture that uses a hyper network to establish image-specific perception rules and then leverages a quality prediction network to flexibly predict image quality. Ou \textit{et al.} \cite{ou2022novel} proposed a novel controllable list-wise rank learning–based IQA method, which is called CLRIQA, that synthesizes authentic distortion–based rank samples and employs a controllable list-wise ranking loss function to produce a robust prediction model. Ma \textit{et al.} \cite{ma2023forgetting} proposed a scalable incremental learning framework (SILF) that enables sequential blind IQA across multiple evaluation tasks under limited memory constraints. In recent years, researchers have begun evaluating the quality of image observations within intelligent control scenarios. For example, the study conducted by Zhang \textit{et al.} \cite{zhang2024embodied} introduced an embodied IQA framework for robotic intelligence, which synergistically integrates task-driven perceptual fidelity metrics with a novel embodied preference database (EPD) of 5,000 reference and distorted image annotations to uncover fundamental disparities between HVS and RVS.

However, mainstream IQA methods solely aim to predict the visual quality perceived by HVS \cite{shen2019no, jiang2020blind, ou2022novel, tian2022vsoiqe, song2022medical, ou2023troubleshooting, zhu2023blind, yue2024perceptual, gao2024blind, ou2024refining, ou2024clib}, and the existing embodied IQA framework \cite{zhang2024embodied} for RVS predominantly concentrates on image observations generated within simulation environments and remains underexplored in the context of concrete intelligent control applications. In addition, the embodied IQA framework \cite{zhang2024embodied} is confined to single-perspective image observation, whereas in practical applications, robots typically integrate image observations from multiple viewpoints to inform their decision‑making processes. Therefore, to bridge these gaps, we first establish the MyRMB dataset that quantifies the quality of real-world-based multi-view image observations through robotic behaviors. Then, we develop an objective IQA method to evaluate the image quality. Moreover, our proposed IQA approach is comprehensively investigated in specific robotic control applications.

\section{Cable Routing Task}
Cable routing has long been regarded as a benchmark problem for assessing robotic dexterity and sequential decision-making, owing to the deformable nature of the medium and the necessity for precise, multistage manipulation \cite{luo2024multi}. Achieving robust, repeatable success in this task is a critical prerequisite for the widespread deployment of robots in manufacturing and maintenance operations, where efficient and reliable cable installation under stringent spatial and tolerance constraints is paramount \cite{luo2024multi}.

As configured in the study conducted by Luo \textit{et al.} \cite{luo2024multi}, the robot, which is controlled by a multistage controller, is required to route a cable through three individual clips by systematically executing a sequence of skills. In the high-level skill selection task, the high-level policy of the robot needs to select a discrete skill. Once a skill is selected, a corresponding low-level policy will attempt to accomplish it over the subsequent period of time. In the low-level routing task, the low-level policy needs to generate a sequence of concrete continuous actions to complete the skill. For the purpose of training such a multistage controller, they built two IL datasets that were collected via human experts.

Despite attaining the expected performance on the cable routing tasks, their methodology fails to account for potential distortions in image observations under practical deployment environments. Thus, building upon their foundational datasets, we extend image observations into distorted scenarios. Additionally, we propose a framework in which an IQA methodology is combined with a confidence‑based learning mechanism to significantly improve decision-making precision in IL.

\section{The MyRMB Dataset}
In Table \ref{Summary of popular datasets for robotics and image quality assessment}, we summarize a number of well-known datasets for both robotics and IQA areas. It can be seen that there is an urgent need for a dataset enabling robotic IL in distorted environments while revealing the impact of image observation quality on RVS. To this end, the MyRMB dataset is developed. With this dataset, the quality of image observations in robotics can now be quantitatively evaluated, thereby providing a foundation for the development of our confidence-aware IL methodology.

\begin{table*}[tbp]
	\centering
	\caption{Summary of popular datasets for robotics and image quality assessment}
	\label{Summary of popular datasets for robotics and image quality assessment}
    \resizebox{\linewidth}{!}{
	\begin{tabular}{cllll}
		\toprule
		\textbf{Domains} & \textbf{Datasets} & \textbf{Venues} & \textbf{Amount} & \textbf{Tasks} \\
		\midrule
		\multirow{6}*{Robotics} & Dataset released by Finn \textit{et al.} \cite{finn2016unsupervised} & NeurIPS 2016 & 59,000 interactions & Pushing \\
            & Dataset released by Levine \textit{et al.} \cite{levine2018learning} & IJRR 2018 & over 800,000 grasp attempts & Grasping \\
            & RoboNet \cite{dasari2020robonet} & CoRL 2019 & 15 million video frames & Generalizable robotic learning\\
            & BridgeData V2 \cite{walke2023bridgedata} & CoRL 2023 & 60,096 trajectories & Scalable robot learning \\
            & Interactive Language \cite{lynch2023interactive} & RA-L 2023 & 600,000 trajectories & Building interactive, real-time, natural language-instructable robots \\
            & Datasets released by Luo \textit{et al.} \cite{luo2024multi} & T-RO 2024 & 1,699 trajectories & Multistage cable routing \\
            \midrule
            \multirow{5}*{IQA} & MLIVE \cite{jayaraman2012objective} & ACSCC 2012 & 420 images & Objective IQA for two types of multiply distorted images \\
            & MDID2013 \cite{gu2014hybrid} & TBC 2014 & 336 images & IQA with multiple distortion types \\
            & TID2013 \cite{ponomarenko2015image} & SPIC 2015 & 3,025 images & Evaluation of full-reference visual quality assessment measures \\
            & KADID-10k \cite{lin2019kadid} & QoMEX 2019 & 10,206 images & Deep learning for IQA \\
            & Degraded-Reference IQA \cite{athar2023degraded} & TIP 2023 & 32,946 images per version & Development of robust machine learning based IQA models \\
        \midrule
            \multirow{2}*{IQA for Robotics} & EPD \cite{zhang2024embodied} & arXiv 2024 & 5,000 images & Exploration of IQA in embodied artificial intelligence \\
            & MyRMB (ours) & -- & 56,992 samples & Cable routing in scenarios involving image distortion \\
		\bottomrule
	\end{tabular}}
\end{table*}

\begin{figure}[tbp]
	\centering
	\captionsetup[subfloat]{labelfont=scriptsize,textfont=scriptsize}
	\subfloat[RF: ED=0.0000]{
		\begin{minipage}[b]{0.45\linewidth}
			\includegraphics[width=1\linewidth]{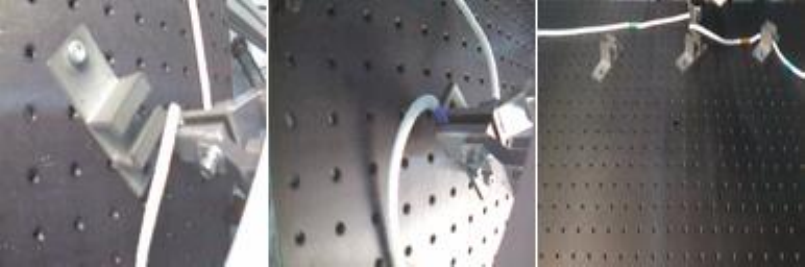}
		\end{minipage}
		\label{hrf}
	}
	\subfloat[AGN: ED=1.4737]{
		\begin{minipage}[b]{0.45\linewidth}
			\includegraphics[width=1\linewidth]{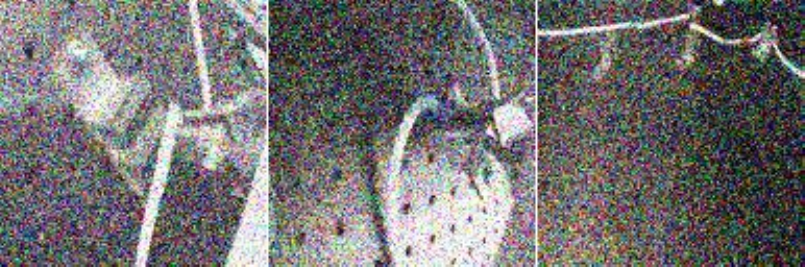}
		\end{minipage}
		\label{hagn}
	}\\
	\subfloat[JP: ED=0.1253]{
		\begin{minipage}[b]{0.45\linewidth}
			\includegraphics[width=1\linewidth]{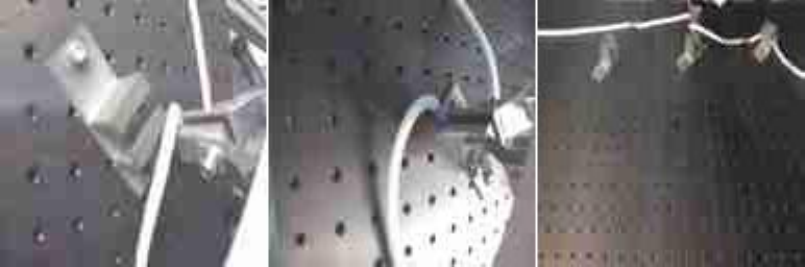}
		\end{minipage}
		\label{hjp}
	}
	\subfloat[GB: ED=1.1525]{
		\begin{minipage}[b]{0.45\linewidth}
			\includegraphics[width=1\linewidth]{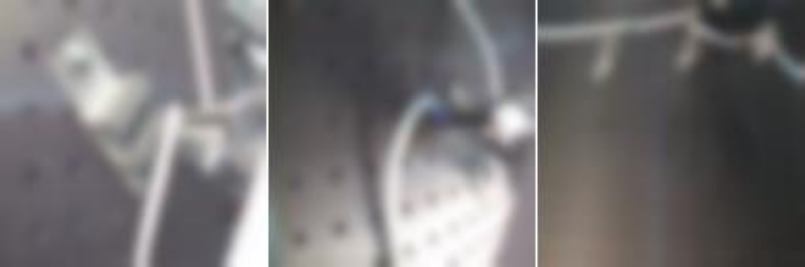}
		\end{minipage}
		\label{hgb}
	}\\
	\subfloat[MB: ED=0.4865]{
		\begin{minipage}[b]{0.45\linewidth}
			\includegraphics[width=1\linewidth]{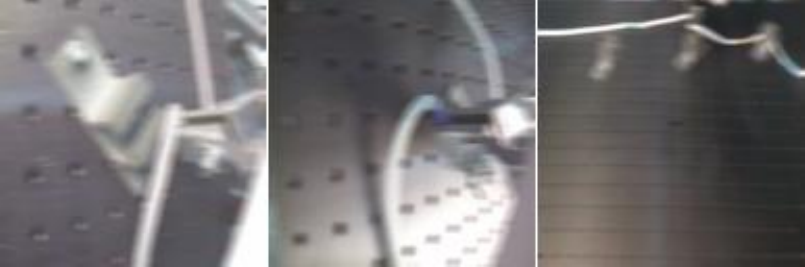}
		\end{minipage}
		\label{hmb}
	}
	\subfloat[MD: ED=2.4023]{
		\begin{minipage}[b]{0.45\linewidth}
			\includegraphics[width=1\linewidth]{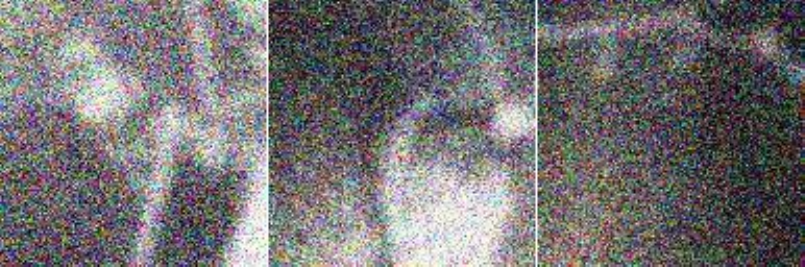}
		\end{minipage}
		\label{hmd}
	}
	\caption{Examples of image observations for the high-level skill selection task in our dataset. The image observations include two wrist camera observations and a side camera observation. The reference set of image observations (RF) presented in this example is selected randomly from our dataset. The quality scores, which are Euclidean distance (ED) values, are displayed. Here, the distortion types include additive Gaussian noise (AGN), JPEG compression (JP), Gaussian blur (GB), motion blur (MB), and multiple distortion (MD).}
	\label{highlevel_dataset}
\end{figure}

\begin{figure}[tbp]
	\centering
	\captionsetup[subfloat]{labelfont=scriptsize,textfont=scriptsize,justification=centering}
	\subfloat[RF: ED=0.0000]{
		\begin{minipage}[b]{0.3\linewidth}
                \centering
			\includegraphics[width=0.95\linewidth]{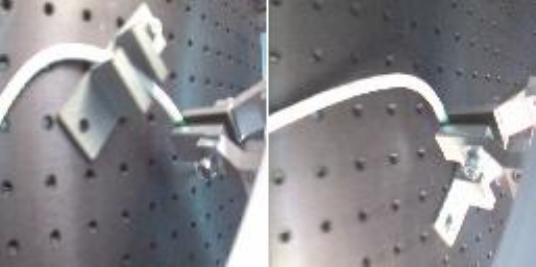}
		\end{minipage}
		\label{lrf}
	}
	\subfloat[AGN: ED=0.7628]{
		\begin{minipage}[b]{0.3\linewidth}
                \centering
			\includegraphics[width=0.95\linewidth]{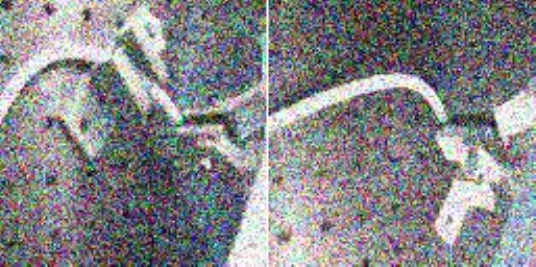}
		\end{minipage}
		\label{lagn}
	}
	\subfloat[JP: ED=0.1474]{
		\begin{minipage}[b]{0.3\linewidth}
                \centering
			\includegraphics[width=0.95\linewidth]{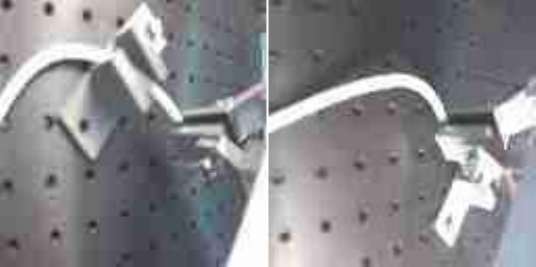}
		\end{minipage}
		\label{ljp}
	}\\
	\subfloat[GB: ED=0.6453]{
		\begin{minipage}[b]{0.3\linewidth}
                \centering
			\includegraphics[width=0.95\linewidth]{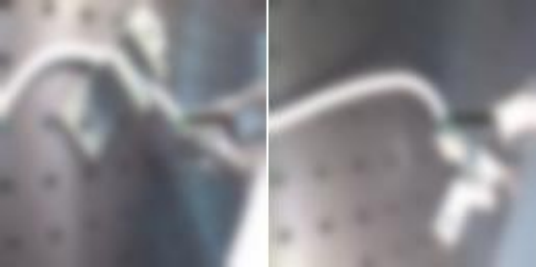}
		\end{minipage}
		\label{lgb}
	}
	\subfloat[MB: ED=0.1517]{
		\begin{minipage}[b]{0.3\linewidth}
                \centering
			\includegraphics[width=0.95\linewidth]{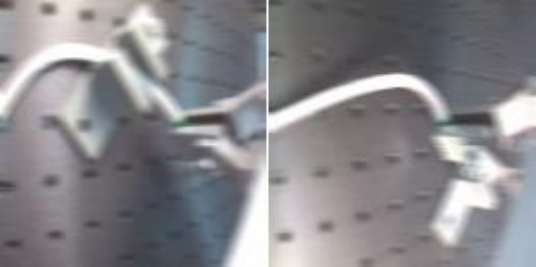}
		\end{minipage}
		\label{lmb}
	}
	\subfloat[MD: ED=0.7236]{
		\begin{minipage}[b]{0.3\linewidth}
                \centering
			\includegraphics[width=0.95\linewidth]{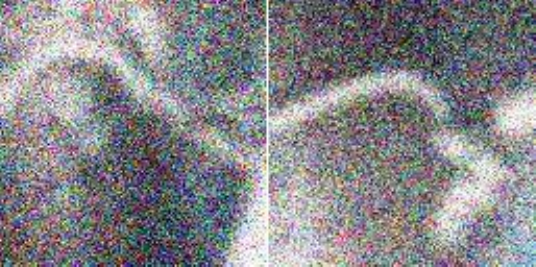}
		\end{minipage}
		\label{lmd}
	}
	\caption{Examples of image observations for the low-level cable routing task in our dataset. Unlike the image observations for the high-level skill selection task, the low-level cable routing task only involves two wrist camera observations.}
	\label{lowlevel_dataset}
\end{figure}

\subsection{Image Collection}
Our MyRMB dataset involves two robotic control tasks, which are the high-level skill selection task and the low-level cable routing task. As presented in Table \ref{Information of our new dataset}, each sample leveraged for high-level behaviors in our dataset consists of two robot wrist camera observations, a side camera observation, a robot's end-effector pose, a six-step history of executed skills, a quality score (\textit{i.e.}, the Euclidean distance value), and a ground-truth discrete skill. In comparison, only two robot wrist camera observations, a robot's end-effector pose, a quality score, and a ground-truth continuous action are included in each sample utilized for low-level behaviors. In the MyRMB dataset, image observations captured from multiple viewpoints correspond to a single quality score, whereas in typical IQA tasks, each image is associated with an individual quality score. All camera observations are of the same size of 128$\times$128 pixels. The reference samples are derived from the robotic IL datasets released by Luo \textit{et al.} \cite{luo2024multi}. 1,192 reference sets of image observations for high-level behaviors and 1,000 reference sets for low-level behaviors are contained. As presented in Fig. \ref{highlevel_dataset} and Fig. \ref{lowlevel_dataset}, there are five typical distortion types included in this dataset, which are 1) additive Gaussian noise, 2) Gaussian blur, 3) motion blur, 4) JPEG compression, and 5) multiple distortion caused by all above types, and five distortion levels are introduced for each distortion type. The distortion types are designed for robotic real-world applications, simulating the distortion situations encountered by robot cameras in reality: 1) the additive Gaussian noise distortion is commonly encountered in low-light imaging conditions or in images captured by sensors with high electronic noise; 2) the Gaussian blur occurs due to defocusing or lens imperfections and is often seen in photography and videography when the camera is not properly focused; 3) the motion blur arises from the relative motion between the camera and the subject during exposure; 4) the JPEG compression distortion is frequently observed in images that have been compressed to reduce file size for storage or transmission; and 5) the multiple distortion occurs when an image is subjected to more than one type of degradation simultaneously. As a result, 29,800 distorted samples (1,192 reference samples $\times$ 5 distortion types $\times$ 5 distortion levels) are created for the high-level decision-making policy, and 25,000 distorted samples (1,000 reference samples $\times$ 5 distortion types $\times$ 5 distortion levels) are made for the low-level.

\begin{table*}[t]
	\centering
	\caption{Information of the MyRMB dataset. This dataset facilitates the training of robots to operate in scenarios with distortions. Moreover, it makes it possible to assess the quality of image observations in robotics, thereby offering valuable image quality information that can be utilized in our confidence-aware IL approach.}
	\label{Information of our new dataset}
	\begin{tabular}{l|l|l}
		\toprule
		\textbf{Items} & \textbf{Information for High-Level Skill Selection Task} & \textbf{Information for Low-Level Routing Task} \\
		\midrule
        Reference Sample Size & 1,192 & 1,000 \\
        \midrule
        Distorted Sample Size & 29,800 & 25,000 \\
        \midrule
        Content of Samples & Wrist Camera Observations, & Wrist Camera Observations, \\
         & Side Camera Observations, & End-Effector Poses, \\
         & End-Effector Poses, & Quality Scores \\
         & Skill Histories, & Ground-Truth Continuous Actions\\
         & Quality Scores & \\
         & Ground-Truth Discrete Skills & \\
        \midrule
        Distortion Types & \multicolumn{2}{c}{Additive Gaussian Noise, Gaussian Blur, Motion Blur, JPEG Compression, Multiple Distortion} \\
        \midrule
        Number of Distortion Levels & \multicolumn{2}{c}{5} \\
        \midrule
        Pixel Dimension of Image Observations & \multicolumn{2}{c}{128$\times$128} \\
		\bottomrule
	\end{tabular}
\end{table*}

\begin{figure}[tbp]
	\centering
	\captionsetup[subfloat]{labelfont=scriptsize,textfont=scriptsize}
	\subfloat[Prediction Accuracy]{
		\begin{minipage}[b]{0.4825\linewidth}
			\includegraphics[width=1\linewidth]{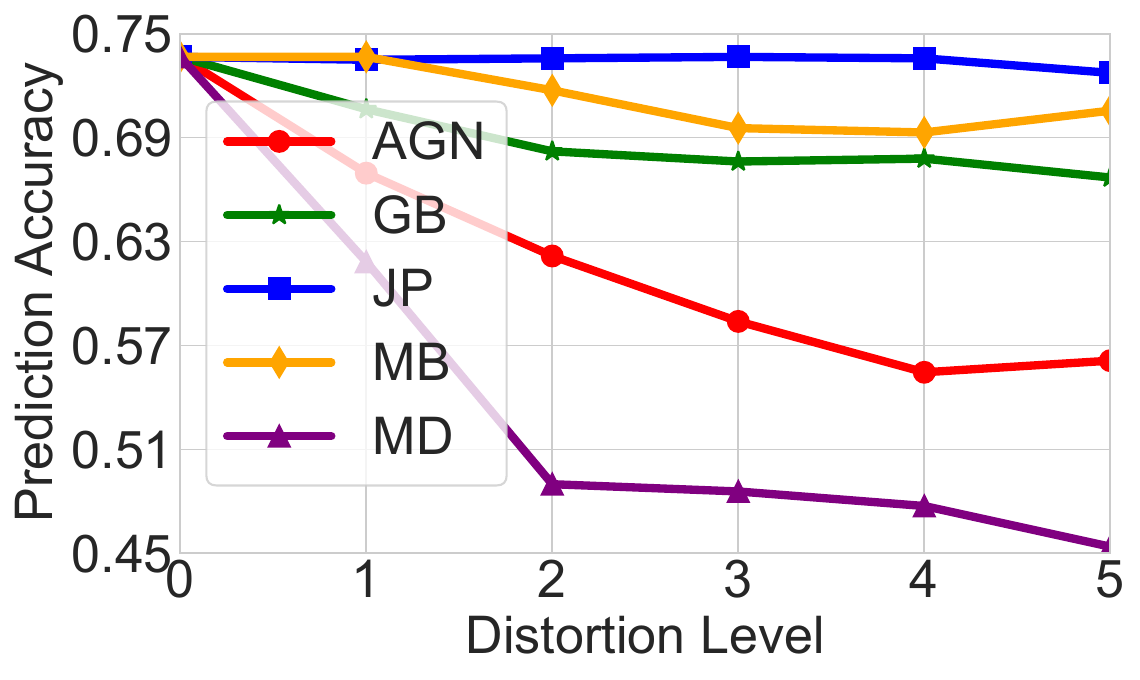}
		\end{minipage}
		\label{haccuracy}
	}
	\subfloat[Euclidean Distance]{
		\begin{minipage}[b]{0.482\linewidth}
			\includegraphics[width=1\linewidth]{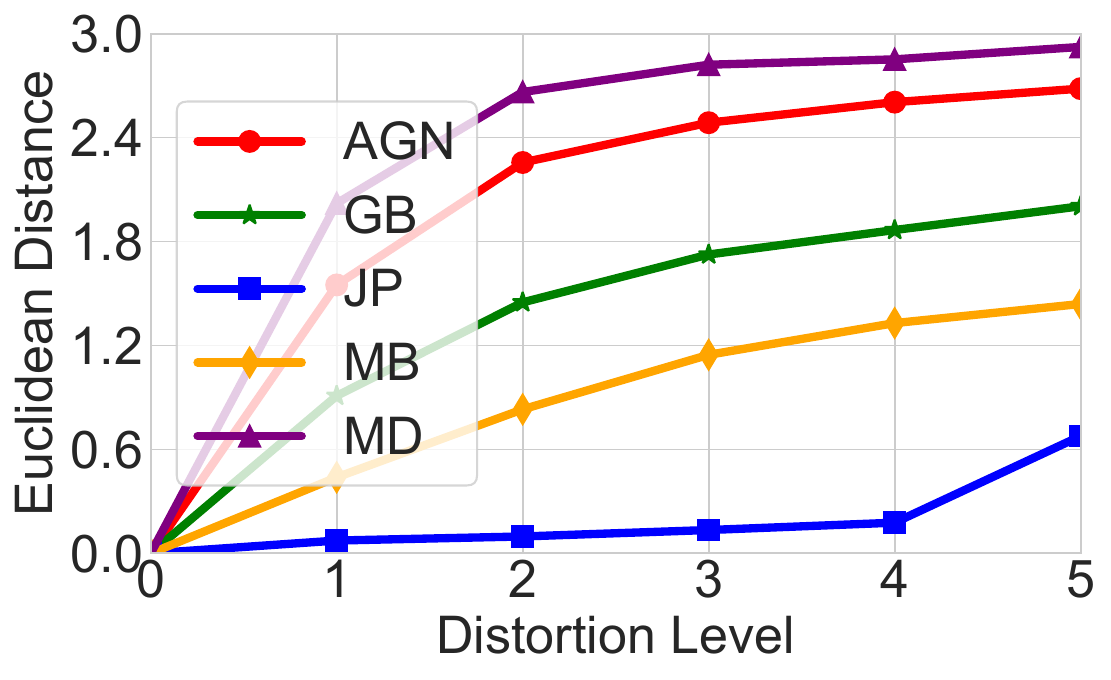}
		\end{minipage}
		\label{hed}
	}
	\caption{Statistical data of samples for the high-level skill selection task. (a) The prediction accuracy of the high-level policy model under diverse distorted conditions. (b) The Euclidean distance values between the embeddings of reference and distorted samples.}
	\label{highlevel_statistics}
\end{figure}

\begin{figure}[tbp]
	\centering
	\captionsetup[subfloat]{labelfont=scriptsize,textfont=scriptsize}
	\subfloat[MAE]{
		\begin{minipage}[b]{0.4825\linewidth}
			\includegraphics[width=1\linewidth]{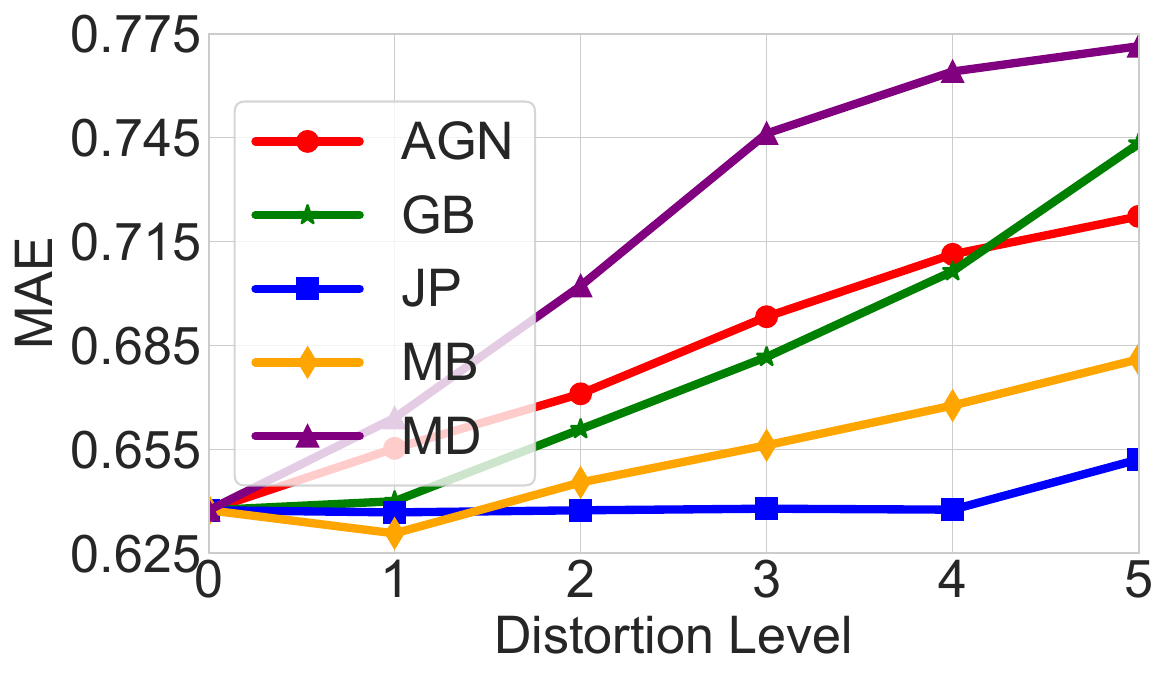}
		\end{minipage}
		\label{lmae}
	}
	\subfloat[Euclidean Distance]{
		\begin{minipage}[b]{0.4825\linewidth}
			\includegraphics[width=1\linewidth]{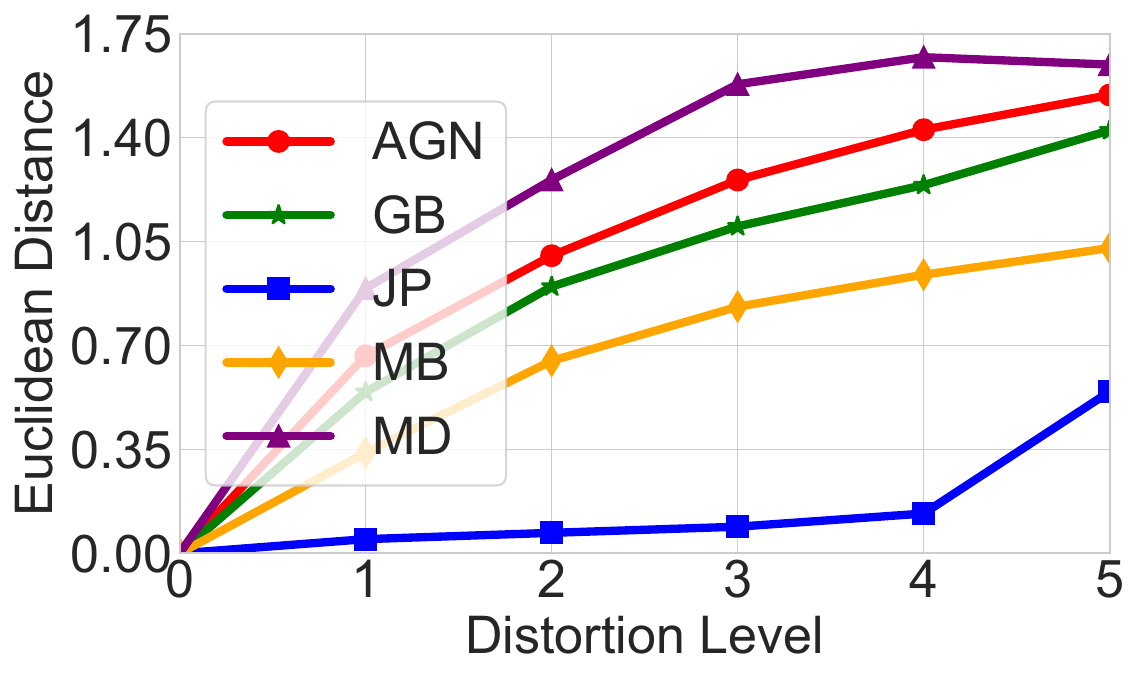}
		\end{minipage}
		\label{led}
	}
	\caption{Statistical data of samples for the low-level cable routing task. (a) The MAE values between predicted low-level actions and ground-truth values under diverse distorted conditions. (b) The Euclidean distance values between the embeddings of reference and distorted samples.}
	\label{lowlevel_statistics}
\end{figure}

\subsection{Quality Score Collection}
The MyRMB dataset also eliminates the need for time-consuming and labor-intensive manual scoring, which is typically required in traditional datasets, for obtaining image quality scores. In this dataset, Euclidean distance values between the embeddings of reference and distorted samples are regarded as quality scores. At first, to investigate the impact of distorted image observations on decision-making tasks, the performance of the policy models proposed by Luo \textit{et al.} \cite{luo2024multi} is evaluated in distorted scenarios. On the high-level skill selection task, we test the prediction accuracy of the high-level policy model. The prediction accuracy $\xi$ is described as $\xi = \frac{i}{n}$, where $i$ represents the number of correct predictions, and $n$ denotes the total number of predictions. As depicted in Fig. \ref{highlevel_statistics}, with the increase in the degree of distortion, the prediction accuracy of the high-level policy model under each type of distortion is essentially declining. The change in Euclidean distance values between the embeddings of reference and distorted samples is roughly inverse to the change in prediction accuracy. On the other hand, to explore the interference of visual distortion with low-level decision-making, the mean absolute error (MAE) values between predicted low-level actions and ground truth values are tested. As shown in Fig. \ref{lowlevel_statistics}, as the degree of distortion increases, the MAE values gradually increase, which is similar to the change in Euclidean distance values. Since it is revealed that the Euclidean distance values between distorted samples' and reference samples' embeddings can be regarded as measures to characterize the performance of robots in human-level control tasks, they are collected as quality scores via the policy models proposed by Luo \textit{et al.} \cite{luo2024multi}.

\section{The Proposed Robotic Imitation Learning Framework}
In the proposed framework, the integration of an IQA module with a confidence-based learning mechanism is used to improve the performance of the decision-making module under visual signal distortions. This section provides a detailed description of the proposed robotic IL framework. 

\begin{figure*}[tbp]
	\centering
	\includegraphics[width=\linewidth]{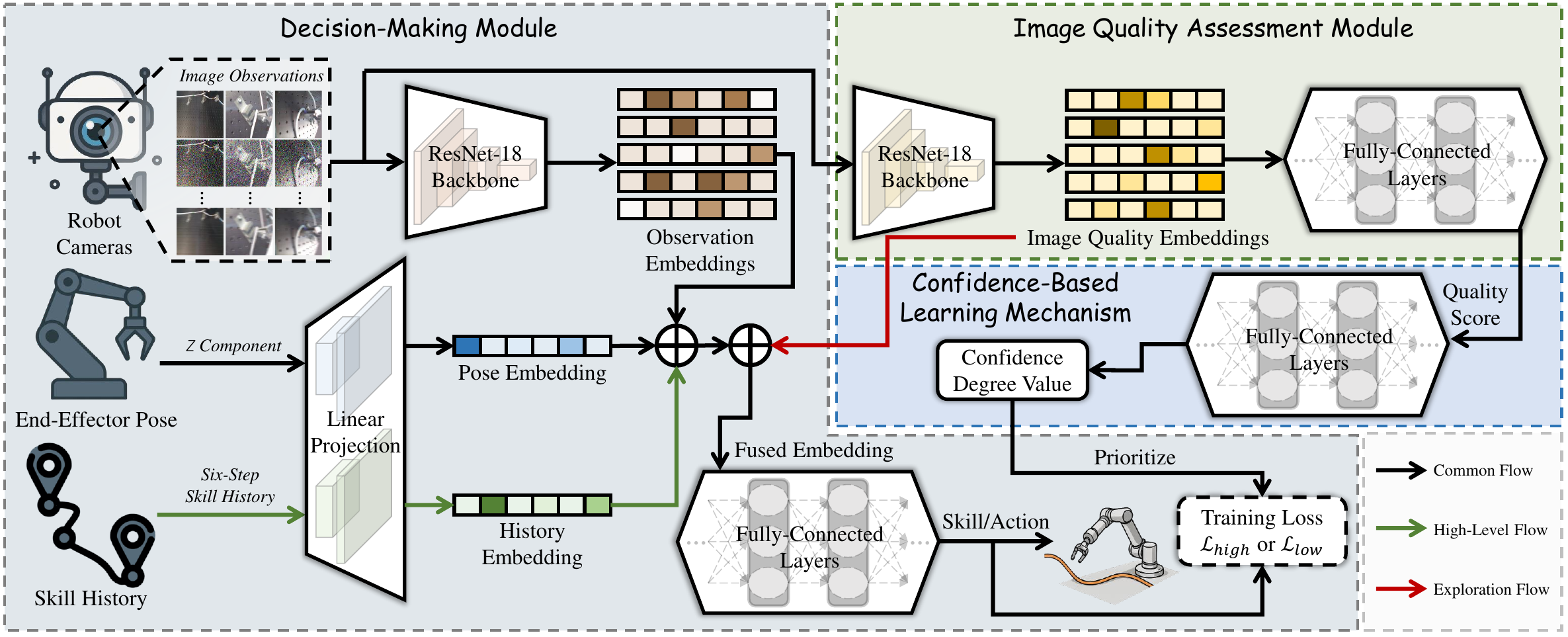}
	\caption{Overview of the proposed IL framework. \textit{Common Flow} denotes its presence in both the high-level skill selection task and the low-level routing task. \textit{High-Level Flow} indicates that it appears exclusively in the high-level skill selection task. \textit{Exploration Flow} refers to its role in further investigation for improving decision-making performance, which will be discussed in Section \ref{Exploration for further enhancing decision-making performance}.}
	\label{overview_il}
\end{figure*}

\subsection{Framework Overview}
As depicted in Fig. \ref{overview_il}, the proposed framework comprises three principal components, including an IQA module, a confidence-based learning mechanism, and an decision-making module. Herein, the IQA module primarily extracts image quality information from image observations via an IQA model. The confidence-based learning mechanism utilizes a quality-aware confidence degree generator model to generate confidence degrees based on image quality priors, thereby directing the training of the decision-making module. The decision-making module subsequently leverages these confidence-guided samples to progressively refine the capability of decision-making models.

\subsection{Image Quality Assessment Model}
In the context of autonomous robotic decision-making tasks, robot cameras typically capture visual observations from multiple perspectives. Consequently, the multi-perspective visual observations are associated with a singular quality score. Mathematically, the IQA model can be expressed as $\hat{q}=f_{\phi}(I_1, I_2,...,I_n)$, where $f_{\phi}$ denotes the IQA model with network parameters $\phi$, and $I_i$ is the $i$-th image observation. In this study, the robot acquires visual observations from three distinct viewpoints in the high-level skill selection task, including two streams from wrist-mounted cameras (denoted $I_{1}$ and $I_{2}$) and a stream from a side camera (denoted $I_{3}$). In the low-level routing task, visual observations are captured from two separate viewpoints, consisting of two wrist camera observations (denoted $I_{1}$ and $I_{2}$). Therefore, the whole procedure of the IQA model can be described as:
\begin{equation}
	\hat{q}=
    \begin{cases}
    f_{\phi}(I_1, I_2, I_3), & \quad \text{if high-level task}, \\
    f_{\phi}(I_1, I_2), & \quad \text{if low-level task}.
    \end{cases}
    \label{iqa_hat_q}
\end{equation}

In the IQA model, multi-view image observations are processed through the corresponding ResNet-18 networks \cite{he2016deep} to extract image features. These features are subsequently fused and fed into fully-connected layers to derive the quality score. To optimize this IQA model such that the predicted quality score $\hat{q}$ closely approximates the ground-truth value $q$, the regression loss $\mathcal{L}_{IQA}$ is minimized:
\begin{equation}
\begin{split}
    \mathcal{L}_{IQA} &= ||\hat{q}-q||^2_2 \\
    &=
    \begin{cases}
    ||f_{\phi}(I_1, I_2, I_3)-q||^2_2, & \text{if high-level task}, \\
    ||f_{\phi}(I_1, I_2)-q||^2_2, & \text{if low-level task}.
    \end{cases}
\end{split}
\label{iqa_loss}
\end{equation}

\subsection{Quality-Aware Confidence Degree Generator}
The quality-aware confidence degree generator is parameterized by fully-connected layers. Its function is to map quality scores predicted by the IQA model to confidence degree values. The range of confidence degree values is $(0,1)$. Once a set of image observations is processed by the IQA model, the resulting image quality score is passed through the generator model, yielding a Gaussian distribution as the confidence probability distribution. Subsequently, a value sampled from this probability distribution serves as the confidence degree value that implies the challenging level of this training sample for the decision-making model. 

RL is leveraged to optimize the generator model. This setting is motivated by two principal considerations: 1) our investigation reveals that the performance of the decision-making model does not necessarily improve with increased image quality, which will be discussed in Section \ref{sec Decision-making performance across different image quality ranges}; and 2) the set of challenging samples for a decision-making model may shift over successive training iterations. As a consequence, an RL-based optimization strategy can be an excellent tool for flexibly managing challenging samples across heterogeneous image quality levels and dynamically capturing them in different training iterations.

As the generator provides the confidence degree value based on the predicted quality score, the predicted quality score $\hat{q}$ and the confidence degree value $c$ can be treated as a state and an action, respectively, in the RL formulation. Hence, the generator model $g$ with the network parameters $\theta$ can be formulated as $c \sim g_{\theta}(c|\hat{q})$. Combined with (\ref{iqa_hat_q}), the procedure of sampling a confidence degree value can be described as:
\begin{equation}
	c \sim g_{\theta}(c|\hat{q})=
    \begin{cases}
    g_{\theta}(c|f_{\phi}(I_1, I_2, I_3)), & \quad \text{if high-level task}, \\
    g_{\theta}(c|f_{\phi}(I_1, I_2)), & \quad \text{if low-level task}.
    \end{cases}
\end{equation}
Once the confidence degree value is sampled, a reward signal is obtained to evaluate how effectively this confidence degree reflects the relative difficulty level of the training sample for the decision-making model. When a training sample is challenging for the decision-making model to learn, the larger the confidence degree value generated by the generator model, the higher the reward value. The reward function $r$ is defined as:
\begin{equation}
\begin{split}
    r &= -c\log(p) \\
    &= -c\log(\pi(a^{*}|\cdot)),
\end{split}
\label{fr}
\end{equation}
where $p$ represents the probability of executing the ground-truth action, and $\pi(a^{*}|\cdot)$ denotes the probability of executing the ground-truth action $a^{*}$ under the policy function $\pi$ of the decision-making model. Then, based on (\ref{fr}), the reward is normalized as:
\begin{equation}
\begin{split}
	r_{norm} &= \frac{r-\mu_r}{\sigma_r} \\
    &= -\frac{c\log(\pi(a^{*}|\cdot))+\mu_r}{\sigma_r},
\end{split}
\end{equation}
where $r_{norm}$ is the normalized reward, and $\mu_r$ and $\sigma_r$ represent the mean value and the standard deviation of rewards for the batch of samples, respectively. The generator model is trained to achieve the normalized reward as high as possible. To this end, the policy gradient theory \cite{williams1992simple, sutton1999policy} is used to design the loss function $\mathcal{L}_{RL}$:
\begin{equation}
	\mathcal{L}_{RL}=-\log(g_{\theta}(c|\hat{q}))\cdot r_{norm}.
\end{equation}
Based on this generator network, the generated confidence values prioritize training samples according to their difficulty for the decision-making network to learn, thus facilitating the training of the decision-making model. The decision-making model will put more effort into learning challenging samples.

\subsection{High-Level Decision-Making Model}
\label{High-Level Decision-Making Model}
In this proposed framework, deep neural networks are employed for robotic decision-making, adopting the network architectures introduced by Luo \textit{et al.} \cite{luo2024multi}. The network model used for decision-making on the high-level skill selection task is referred to as the high-level decision-making model. In this subsection, the high-level decision-making model will be described in detail. 

The high-level decision-making model takes two wrist camera observations (\textit{i.e.}, $I_{1}$ and $I_{2}$), a side camera observation $I_{3}$, an end-effector pose vector $z$, and a history of skills $h$ as input and outputs a selected skill $\hat{u}$ at the given state. Due to the fact that the high-level policy is required to be abstract and long-term, image observations from both global and local perspectives need to be fed into the high-level decision-making model. Thus, image observations from a side perspective and two wrist perspectives are first input into the corresponding convolutional neural networks (CNNs), \textit{i.e.}, ResNet-18 networks \cite{he2016deep} with group normalization \cite{wu2018group}, to obtain three embedding vectors, respectively. Subsequently, these embedding vectors are processed by an averaged fusion to get an observation embedding. Meanwhile, since the $z$ component of the end-effector pose and the skill history also play a crucial role in enhancing the decision-making performance of robots, they are passed through corresponding layers to obtain embedding vectors. Here, a fully-connected layer processes the $z$-coordinate to obtain a higher-dimension pose embedding vector. At the same time, the six-step skill history is input to a learned word embedding layer \cite{mikolov2013distributed} to get a history embedding vector. The skills that may be involved in the skill history include \textit{Route}, \textit{Pickup}, \textit{Perturb}, \textit{Go Next}, and \textit{Empty Padding}. The \textit{Route} skill aims to insert the held cable into the nearest clip. The \textit{Pickup} skill is employed to grasp the cable at a specific location and retain it within the gripper. The \textit{Perturb} skill enables the reconfiguration of the cable into a different shape. The \textit{Go Next} skill is used to route the cable into the following clip. The \textit{Empty Padding} is leveraged to fill the empty slots if the current number of executed skills is less than six. Finally, the observation embedding vector, the pose embedding vector, and the history embedding vector are concatenated together and passed through fully-connected layers to get a discrete skill $\hat{u}$. 

The high-level skill selection task can be regarded as a four-class classification problem within the realm of supervised learning. Therefore, the whole procedure of the high-level decision-making model can be described as $\hat{u}\sim\pi^{h}_{\omega}(\hat{u}|I_{1},I_{2},I_{3},z,h)$, where $\pi^{h}_{\omega}$ is the high-level policy with the network parameters $\omega$. As it is expected that the dissimilarity between the predicted probability distribution $\pi^{h}_{\omega}(u|I_{1},I_{2},I_{3},z,h)$ and the ground-truth distribution is as small as possible, the loss of the high-level decision-making model $\mathcal{L}_{high}$ is minimized over the training set:
\begin{equation}
	\mathcal{L}_{high}=-c\log(\pi^{h}_{\omega}(u|I_{1},I_{2},I_{3},z,h)),
\end{equation}
where $u$ is the ground-truth selected skill, and $c$ denotes the quality-aware confidence degree value. By using the proposed loss function $\mathcal{L}_{high}$, the decision-making model $\pi^{h}_{\omega}$ is able to place greater emphasis on challenging samples during training, ultimately improving its decision-making performance.

\subsection{Low-Level Decision-Making Model}
\label{Low-Level Decision-Making Model}
The model leveraged to control the robot on the low-level cable routing task is regarded as the low-level decision-making model. In this subsection, we will explain this model in detail. 

The function of the low-level decision-making model is to generate concrete continuous actions based on image observations and sensor values. Since the low-level action planning needs to focus on local information, the inputs to the low-level decision-making model are only two wrist camera observations (\textit{i.e.}, $I_{1}$ and $I_{2}$) and the $z$ component of the end-effector pose. Similar to the high-level decision-making model, the wrist camera observations are fed into corresponding ResNet-18 networks \cite{he2016deep} with group normalization \cite{wu2018group} to get a low-level observation embedding. The $z$ component of the end-effector pose is input to a fully-connected layer to obtain a higher-dimension pose embedding. Next, the low-level observation embedding and the pose embedding are concatenated and passed through fully-connected layers to obtain an action probability distribution. The final continuous action is sampled from this distribution.

The concrete continuous action $\hat{a}$ predicted by the low-level decision-making model can be represented as $\hat{a}\sim\pi^l_{\psi}(\hat{a}|I_{1},I_{2},z)$, where $\pi^l_{\psi}$ denotes the low-level decision-making network model with the parameters $\psi$. To reduce the difference between the predicted action probability distribution $\pi^l_{\psi}(a|I_{1},I_{2},z)$ and the ground-truth distribution, the loss of the low-level decision-making model $\mathcal{L}_{low}$ is minimized:
\begin{equation}
	\mathcal{L}_{low}=-c\log(\pi^l_{\psi}(a|I_{1},I_{2},z)),
\end{equation}
where $a$ represents the ground-truth continuous action. 

\section{Experiments}
\label{Experiments}
In this section, the implementation details of our proposed framework are described. We also represent the experimental results and conduct a thorough analysis.

\subsection{Dataset}
Our new dataset consists of 30,992 samples for robotic behaviors on the high-level skill selection task and 26,000 samples for robotic behaviors on the low-level cable routing task, which are utilized to evaluate our proposed IL framework. 

In this experiment, Euclidean distance values between the embeddings of reference and distorted samples are referred to as quality scores. All quality scores are normalized to a range of $[0,1]$. The normalized quality score $q$ can be represented as follows:
\begin{equation}
    q = \frac{\bar{q}_{max} - \bar{q}}{\bar{q}_{max} - \bar{q}_{min}},
\end{equation}
where $\bar{q}_{max}$ and $\bar{q}_{min}$ are the maximum value and the minimum value of original quality scores, respectively, and $\bar{q}$ denotes the original value of the quality score.

\subsection{Evaluation Measures}
In the evaluation of decision-making models, we perceive classification accuracy as prediction accuracy to evaluate the model performance on the high-level skill selection task, while the MAE between generated low-level continuous actions and ground-truth actions is utilized to gauge the model performance on the low-level routing task. 

For gauging the efficacy of IQA models, Pearson Linear Correlation Coefficient (PLCC), Spearman Rank-order Correlation Coefficient (SRCC), Kendall Rank-order Correlation Coefficient (KRCC), and Root-Mean-Squared Error (RMSE) are employed to evaluate the monotonicity and accuracy of the predictions. 

Both high-level and low-level samples are split according to reference images, with 60\% used for training, 20\% for validating, and 20\% for testing. The network model trained with the best validation loss is utilized to compute the test values.

\subsection{Experimental Settings}
Our models are trained on a single NVIDIA L20 GPU. The dimensions of the pose embedding and the history embedding are set to 64 and 24, respectively. The fully-connected layers in the confidence degree generator model and the IQA model are all configured with a structure of (256, 256, 1), and those in the high-level and low-level decision-making models are all set to (256, 256, 4). The number of training epochs is 100. On the high-level skill selection task, the batch size and the learning rate are set to 128 and 3e-4, respectively. On the low-level routing task, the batch size and the learning rate are 256 and 1e-3, respectively.

\subsection{Results and Analysis}
\subsubsection{Overall performance on decision-making tasks}
The proposed IL framework is trained and evaluated on the MyRMB dataset. Its performance is compared with the baseline method developed by Luo \textit{et al.} \cite{luo2024multi}. To validate the role of quality-aware confidence degree values, the decision-making performance of utilizing quality scores directly as confidence degree values, without the incorporation of the quality-aware confidence degree generator, is evaluated.

\begin{table}[tbp]
	\centering
	\caption{Performance comparison on decision-making tasks. The prediction accuracy and the MAE are evaluated on the high-level skill selection task and the low-level routing task, respectively.}
	\label{Performance comparison on decision-making tasks}
    \resizebox{\linewidth}{!}{
	\begin{tabular}{ccc}
        \toprule
		\textbf{Methods} & \textbf{Prediction Accuracy} & \textbf{MAE} \\
        \midrule
		Baseline \cite{luo2024multi}  & 0.8712 & 0.6733 \\
        Ours (no confidence degree generator) & 0.8910 & 0.6741 \\
        Ours & \textbf{0.8931} & \textbf{0.6502} \\
        \bottomrule
	\end{tabular}}
\end{table}

As listed in Table \ref{Performance comparison on decision-making tasks}, our proposed framework exhibits superior performance on both high-level and low-level tasks. With the augmentation of the quality-aware confidence degree generator, our framework achieves a prediction accuracy of nearly 0.9, which outperforms the baseline \cite{luo2024multi} by 2.5138\% on the high-level skill selection task. On the low-level routing task, the improvement achieved by our framework is more pronounced. The MAE between the actions generated by our framework and the ground-truth values is approximately 0.65, which outperforms the baseline \cite{luo2024multi} by 3.4309\%. When quality scores are adopted directly as confidence degree values, it exhibits a 2.2727\% performance improvement compared to the baseline \cite{luo2024multi} on the high-level task, while no performance enhancement is achieved on the low-level task. On both high-level and low-level tasks, the performance of our framework with the confidence generator outweighs that without the generator. Therefore, the results suggest that the proposed quality-aware confidence degree values are more conducive to maintaining excellent performance in scenarios involving image signal distortion.

\subsubsection{Decision-making performance across different image quality ranges}
\label{sec Decision-making performance across different image quality ranges}
\begin{figure*}[tbp]
	\centering
	\captionsetup[subfloat]{labelfont=scriptsize,textfont=scriptsize}
	\subfloat[Prediction accuracy on the high-level task]{
		\begin{minipage}[b]{0.5\linewidth}
			\includegraphics[width=0.95\linewidth]{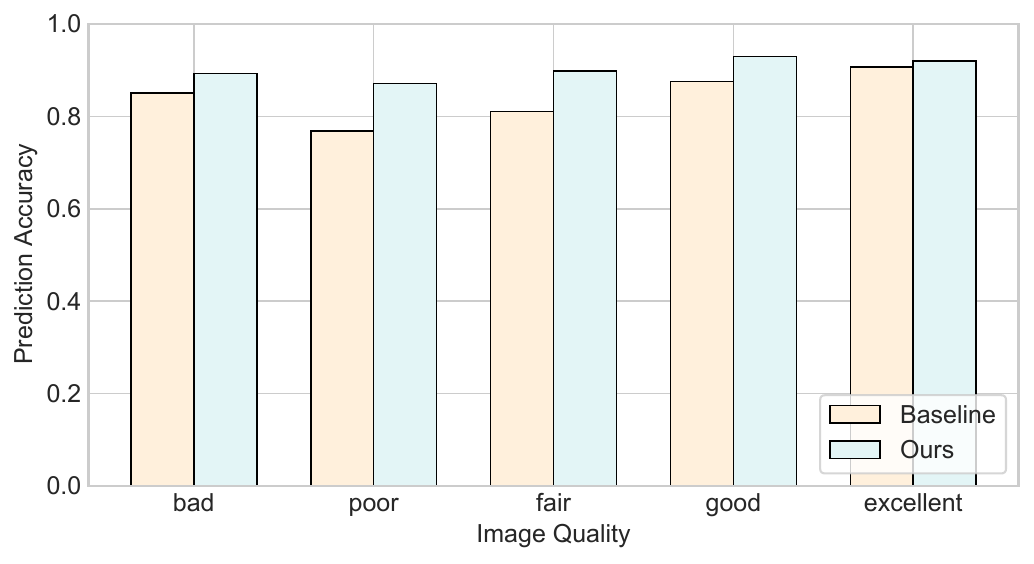}
		\end{minipage}
		\label{h_compare_two}
	}
	\subfloat[MAE on the low-level task]{
		\begin{minipage}[b]{0.5\linewidth}
			\includegraphics[width=0.95\linewidth]{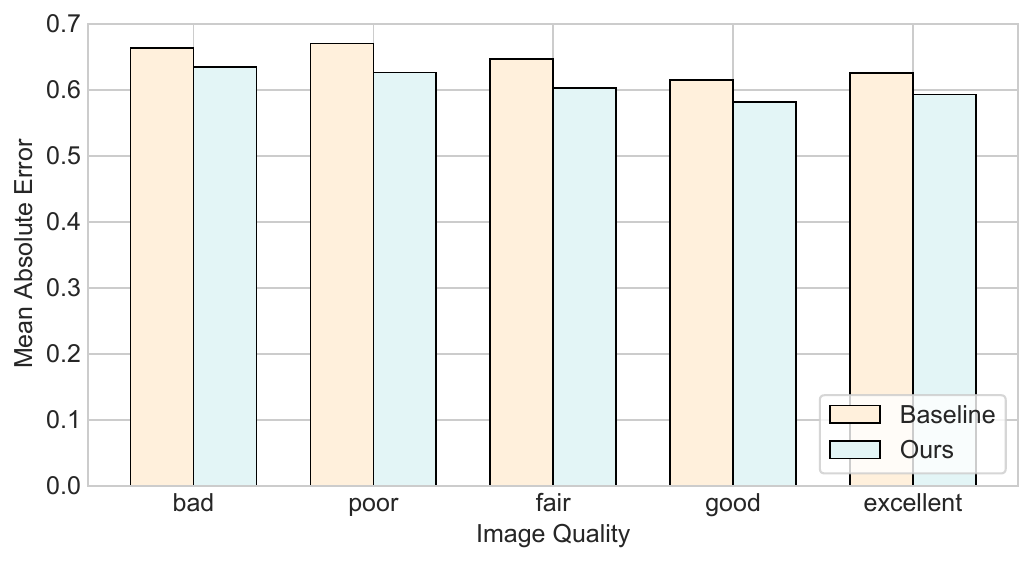}
		\end{minipage}
		\label{l_compare_two}
	}
	\caption{Decision-making performance across different image quality ranges. The samples in the test set are divided into five equal-size groups based on the quality of the image observations, ranging from \textit{bad} to \textit{excellent}.}
	\label{Decision-making performance across different image quality ranges}
\end{figure*}

The decision-making performance across distinct image quality groups is presented in Fig. \ref{Decision-making performance across different image quality ranges}. The results demonstrate that our proposed framework leads to performance improvements across all sample groups. It has a more pronounced effect on low-quality image observation samples, such as those in \textit{poor} and \textit{fair} groups on both high-level and low-level tasks. It is also obvious that our framework exhibits relatively small performance differences across samples with varying image quality compared to the baseline \cite{luo2024multi}.

As the image quality increases, the decision-making performance of both methods exhibits an overall upward trend because richer visual features and structural information provided by higher-quality image observations enable decision-making models to better interpret complex patterns and contextual details. Nevertheless, the performance of the decision-making model does not necessarily improve with the increase in image quality. For instance, as presented in Fig. \ref{Decision-making performance across different image quality ranges}, the performance of the two methods on the samples with \textit{bad} quality is superior to that on the samples with \textit{poor} quality on the high-level task, while on the low-level task, the performance on the samples in \textit{good} group outperforms that on the samples in \textit{excellent} group. This phenomenon suggests that additional details brought by higher-quality image observations may not substantially contribute to task-specific decision-making. Rather than directly using image quality as an indicator of sample difficulty, our quality-aware confidence degree values are designed to dynamically reflect the sample difficulty based on the image quality. Therefore, it enhances the performance of the decision-making model.

\subsubsection{Overall performance evaluation on the IQA model}
The effectiveness of our IQA methodology is also evaluated. We compare our IQA method with five IQA methods, which are the peak signal-to-noise ratio (PSNR), the structural similarity index measure (SSIM) \cite{wang2004image}, the learned perceptual image patch similarity (LPIPS) \cite{zhang2018unreasonable}, the deep image structure and texture similarity (DISTS) \cite{ding2020image}, and CLIPIQA+ \cite{wang2023exploring}. In particular, the PSNR value is normalized as follows:
\begin{equation}
    q_{psnr} = \frac{\bar{q}_{psnr} - \bar{q}^{min}_{psnr}}{\bar{q}^{max}_{psnr} - \bar{q}^{min}_{psnr}},
\end{equation}
where $q_{psnr}$ is the normalized PSNR value, $\bar{q}_{psnr}$ is the original PSNR value, as well as $\bar{q}^{max}_{psnr}$ and $\bar{q}^{min}_{psnr}$ are the maximum PSNR value and the minimum PSNR value, respectively. The normalized LPIPS value $q_{lpips}$ can be calculated as follows:
\begin{equation}
    q_{lpips} = \frac{\bar{q}^{max}_{lpips} - \bar{q}_{lpips}}{\bar{q}^{max}_{lpips} - \bar{q}^{min}_{lpips}},
\end{equation}
where $\bar{q}^{max}_{lpips}$ and $\bar{q}^{min}_{lpips}$ denote the maximum LPIPS value and the minimum LPIPS value, respectively, as well as $\bar{q}_{lpips}$ represents the original LPIPS value. The normalized DISTS value $q_{dists}$ can be described as follows:
\begin{equation}
    q_{dists} = \frac{\bar{q}^{max}_{dists} - \bar{q}_{dists}}{\bar{q}^{max}_{dists} - \bar{q}^{min}_{dists}},
\end{equation}
in which $\bar{q}^{max}_{dists}$ and $\bar{q}^{min}_{dists}$ denote the maximum DISTS value and the minimum DISTS value, respectively, as well as $\bar{q}_{dists}$ is the original DISTS value.

\begin{table}[tbp]
	\centering
	\caption{Overall performance evaluation on IQA via robotic behaviors}
	\label{Overall performance evaluation on IQA via robotic behaviors}
	\begin{tabular}{cccccc}
            \toprule
		\multirow{2}*{\textbf{Levels}} & \multirow{2}*{\textbf{Methods}} & \multicolumn{4}{c}{\textbf{Evaluation Measures}}\\
            \cmidrule{3-6}
		  & & \textbf{PLCC} & \textbf{SRCC} & \textbf{KRCC} & \textbf{RMSE}\\
            \midrule
		\multirow{6}*{High-Level} & PSNR & 0.7469 & 0.8205 & 0.6364 & 0.4228 \\
         & SSIM \cite{wang2004image} & 0.7529 & 0.8256 & 0.6416 & 0.3974 \\
         & LPIPS \cite{zhang2018unreasonable} & 0.7392 & 0.8042 & 0.6198 & 0.3668 \\
         & DISTS \cite{ding2020image} & 0.7030 & 0.7605 & 0.5716 & 0.4500 \\
         & CLIPIQA+ \cite{wang2023exploring} & 0.8346 & 0.8822 & 0.7069 & 0.1021 \\
            \cmidrule{2-6}
         & Ours & \textbf{0.9500} & \textbf{0.9581} & \textbf{0.8293} & \textbf{0.0578} \\
            \midrule
		\multirow{6}*{Low-Level} & PSNR & 0.7538 & 0.7684 & 0.5833 & 0.3958 \\
         & SSIM \cite{wang2004image} & 0.7610 & 0.7917 & 0.6100 & 0.3676 \\
         & LPIPS \cite{zhang2018unreasonable} & 0.7680 & 0.7779 & 0.5935 & 0.3045 \\
         & DISTS \cite{ding2020image} & 0.7778 & 0.7931 & 0.6054 & 0.3604 \\
         & CLIPIQA+ \cite{wang2023exploring} & 0.8592 & 0.8702 & 0.6882 & 0.1002 \\
            \cmidrule{2-6}
         & Ours & \textbf{0.9023} & \textbf{0.9167} & \textbf{0.7546} & \textbf{0.0844} \\
            \bottomrule
	\end{tabular}
\end{table}

The results of PLCC, SRCC, KRCC, and RMSE are reported in Table \ref{Overall performance evaluation on IQA via robotic behaviors}. It can be seen that the performance of our IQA method surpasses that of other methods. Specifically, compared to the best competitor, our IQA model exhibits a 13.827\% enhancement in PLCC, an 8.6035\% increase in SRCC, an improvement of 17.315\% in KRCC, and a reduction of 43.3888\% in RMSE on the high-level skill selection task. On the low-level routing task, it outperforms the best competitor by 5.0163\% PLCC, 5.3436\% SRCC, 9.6484\% KRCC, and 15.7685\% RMSE. Hence, the results indicate that our IQA method can provide the quality-aware confidence degree generator with reliable image quality information. 

\subsubsection{Critical regions of the image for assessing image quality}
\begin{figure}[tbp]
	\centering
	\includegraphics[width=\linewidth]{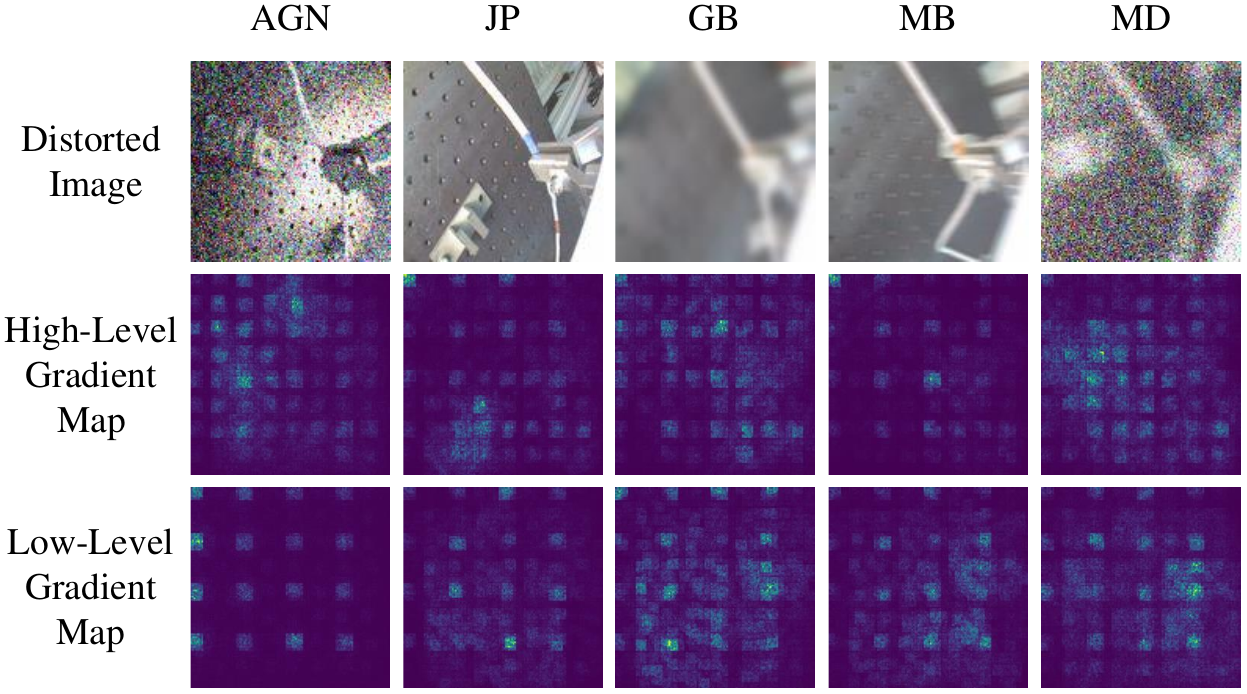}
	\caption{Gradient maps of several distorted images. The distortion types from left to right are additive Gaussian noise (AGN), JPEG compression (JP), Gaussian blur (GB), motion blur (MB), and multiple distortion (MD), respectively.}
	\label{gradient_maps}
\end{figure}

Since Euclidean distance values between the embeddings of reference and distorted samples are used as the image quality scores in this study, image regions that are beneficial for IQA methods to assess image quality can usually be regarded as key areas that influence robotic behaviors in scenarios where visual signals are distorted. It is found that the image areas on which our IQA method focuses differ between the high-level and the low-level tasks. As indicated in Fig. \ref{gradient_maps}, our IQA method prefers to pay attention to clip areas on the high-level task, while it seems to focus on end-effector areas on the low-level task. This implies that the distortion occurring within the clip region of the image observation typically exerts a greater influence on robotic behaviors on the high-level skill selection task, while the distortion within the end-effector region usually has a larger impact on robotic behaviors on the low-level routing task.

\subsubsection{Exploration for further enhancing decision-making performance}
\label{Exploration for further enhancing decision-making performance}
Further enhancement of decision-making performance is explored. As illustrated in Fig. \ref{overview_il}, in this exploration, the image quality embedding extracted by our IQA model is concatenated with other embeddings obtained from the original decision-making model and then jointly fed into the final fully-connected layers, while the other components of our methodology remain unchanged.

As listed in Table \ref{Performance comparison on decision-making tasks with/without the augmentation of image quality embedding}, with the incorporation of image quality embedding, our method shows a slight improvement in performance on the high-level task, with the prediction accuracy exceeding 0.9, while a mild decline in performance is observed on the low-level task. It indicates that the quality-aware confidence degree generator in our methodology may have already conveyed key image quality-related information to the decision-making model, such that additional image quality information does not have a substantial impact on decision-making performance.

\begin{table}[tbp]
	\centering
	\caption{Performance comparison on decision-making tasks with/without the augmentation of image quality embedding. The prediction accuracy and the MAE are evaluated on the high-level skill selection task and the low-level routing task, respectively.}
	\label{Performance comparison on decision-making tasks with/without the augmentation of image quality embedding}
    \resizebox{\linewidth}{!}{
	\begin{tabular}{ccc}
        \toprule
		\textbf{Methods} & \textbf{Prediction Accuracy} & \textbf{MAE} \\
        \midrule
        Without image quality embedding & 0.8931 & \textbf{0.6502} \\
        With image quality embedding & \textbf{0.9002} & 0.6582 \\
        \bottomrule
	\end{tabular}}
\end{table}

\subsubsection{Exploration for further improving IQA precision}
We also explore whether additional prior information (\textit{e.g.}, the end-effector pose $z$ and the skill history $h$) could further improve the precision of the IQA. In this exploration, the decision-making network structures described in Section \ref{High-Level Decision-Making Model} and Section \ref{Low-Level Decision-Making Model} are adopted as IQA network models for the high-level skill selection and the low-level cable routing tasks, respectively, with the sole modification that quality score outputs replace their original skill/action outputs. Therefore, the IQA model can be formulated as follows:
\begin{equation}
	\hat{q}=
    \begin{cases}
    f_{\phi}(I_1, I_2, I_3, z, h), & \quad \text{if high-level task}, \\
    f_{\phi}(I_1, I_2, z), & \quad \text{if low-level task}.
    \end{cases}
\end{equation}
Correspondingly, based on (\ref{iqa_loss}), the regression loss $\mathcal{L}_{IQA}$ becomes:
\begin{equation}
	\mathcal{L}_{IQA}=
    \begin{cases}
    ||f_{\phi}(I_1, I_2, I_3, z, h)-q||^2_2, & \text{if high-level task}, \\
    ||f_{\phi}(I_1, I_2, z)-q||^2_2, & \text{if low-level task}.
    \end{cases}
    \label{iqa_loss_plus}
\end{equation}
In this way, in addition to the image observation information, the end-effector pose and the skill history can also contribute to IQA predictions.

\begin{table}[tbp]
	\centering
	\caption{Performance comparison between image-only and multi-modal IQA models}
	\label{Performance comparison between image-only and multi-modal IQA models}
	\begin{tabular}{cccccc}
		\toprule
		\multirow{2}*{\textbf{Levels}} & \multirow{2}*{\textbf{Methods}} & \multicolumn{4}{c}{\textbf{Evaluation Measures}}\\
            \cmidrule{3-6}
		  & & \textbf{PLCC} & \textbf{SRCC} & \textbf{KRCC} & \textbf{RMSE}\\
            \midrule
		\multirow{2}*{High-Level} & Image-only & 0.9500 & \textbf{0.9581} & \textbf{0.8293} & 0.0578 \\
         & Multi-modal & \textbf{0.9515} & \textbf{0.9581} & 0.8290 & \textbf{0.0571} \\
            \midrule
		\multirow{2}*{Low-Level} & Image-only & 0.9023 & 0.9167 & 0.7546 & 0.0844 \\
         & Multi-modal & \textbf{0.9054} & \textbf{0.9189} & \textbf{0.7577} & \textbf{0.0832} \\
            \bottomrule
	\end{tabular}
\end{table}

In Table \ref{Performance comparison between image-only and multi-modal IQA models}, the performance of the image-only and multi-model IQA models is compared. In this table, the IQA model that receives only image observations as input is designated as \textit{image-only}, whereas the IQA model that incorporates additional information (\textit{e.g.}, the end-effector pose $z$ and the skill history $h$) is designated as \textit{multi-modal}. As listed in Table \ref{Performance comparison between image-only and multi-modal IQA models}, incorporating multi-modal inputs into IQA models can only lead to an insignificant increment in performance. The multi-modal IQA model yields improvements of 0.1579\% in PLCC and 1.2111\% in RMSE on the high-level skill selection task. Within the realm of the low-level routing task, it has contributed to an enhancement of 0.3436\% in PLCC, 0.2400\% in SRCC, 0.4108\% in KRCC, and 1.4218\% in RMSE. These findings suggest that while multi-modal information contributes to enhancing the predictive accuracy of IQA models, its impact remains limited, as the image data alone already enables the IQA models to achieve near-optimal performance.

\section{Conclusion}
\label{Conclusions}
Previous studies have not proposed targeted intelligent control solutions for robotic systems in scenarios involving image signal distortion, nor have they fully utilized image quality information to further enhance the performance of intelligent control methods. To this end, this study introduces a novel robotic IL framework that uses a combination of an IQA module and a confidence-based learning mechanism to enhance the reliability and effectiveness of autonomous robotic systems under conditions of image signal interference. Meanwhile, we construct the MyRMB dataset followed by an objective IQA method to quantify image quality through robotic manipulation behaviors, which provides a tool for researchers to directly assess the adverse effects of image distortions on RVS. However, while our current evaluation has already offered valuable insights into the development of intelligent control systems, we acknowledge that additional validation in diverse physical environments is important to fully demonstrate the practical viability of the proposed framework. Therefore, future research can extend this framework to various physical deployment scenarios, which may uncover new insights and further enhance the robustness of RVS.

\bibliographystyle{IEEEtran}
\bibliography{mylib}

\end{document}